\documentclass[accepted]{uai2022} 

\usepackage[american]{babel}
\usepackage{times}

\usepackage{amsmath,amsfonts,bm}

\def\eqref#1{equation~\ref{#1}}

\def\1{\bm{1}}

\DeclareMathAlphabet{\mathsfit}{\encodingdefault}{\sfdefault}{m}{sl}
\SetMathAlphabet{\mathsfit}{bold}{\encodingdefault}{\sfdefault}{bx}{n}

\usepackage{amsmath,amsfonts,bm}
\usepackage{natbib} 
\usepackage{mathtools} 
\usepackage{booktabs} 
\usepackage{tikz} 
\usepackage{longtable}

\usepackage{hyperref}
\usepackage{url}

\usepackage{here}
\usepackage{algorithm}
\usepackage{algorithmic}
\usepackage{graphicx}
\graphicspath{{iclr2022/images/} }
\usepackage{subfigure}
\newtheorem{theorem}{Theorem}
\newtheorem{claim}[theorem]{Claim}
\newtheorem{note}[theorem]{Note}
\newtheorem{remark}[theorem]{Remark}

\newcommand{\qed}{ \hfill $\square$ \\ }
\def\Proof{\noindent{\bf Proof.}\quad}

\title{Representing Hierarchical Structure \\
by Using Cone Embedding}

\author[1,2]{\href{mailto:<bigwing7714@gmail.com>}{Daisuke Takehara}{}}
\author[1]{\href{mailto:<kei@math.keio.ac.jp>}{Kei Kobayashi}{}}
\affil[1]{%
    Department of Mathematics\\
    Keio University\\
    Yokohama, Kanagawa, Japan
}
\affil[2]{%
    ALBERT.inc\\
    Shinjuku, Tokyo, Japan\\
}

\begin{document}
\maketitle

\begin{abstract}
     Graph embedding is becoming an important method with applications in various areas, including social networks and knowledge graph completion. In particular, Poincaré embedding has been proposed to capture the hierarchical structure of graphs, and its effectiveness has been reported. However, most of the existing methods have isometric mappings in the embedding space, and the choice of the origin point can be arbitrary. This fact is not desirable when the distance from the origin is used as an indicator of hierarchy, as in the case of Poincaré embedding. In this paper, we propose {\bf cone embedding} , embedding method in a metric cone, which solve these problems, and we gain further benefits: 1) we provide an indicator of hierarchical information that is both geometrically and intuitively natural to interpret, and 2) we can extract the hierarchical structure from a graph embedding output of other methods by learning additional one-dimensional parameters.
\end{abstract}

In recent years, machine learning for graph-structured data has attracted significant attention. A recently developed method is using graph convolutional neural networks~\citep{gcn}, which apply convolutional neural networks to graphs.
In this study, we focus on graph embedding, which has become an important method with applications in social networks~\citep{social},  knowledge graph completion~\citep{bordes2013translating}, and other fields. In particular, various methods of embedding in non-Euclidean spaces have been proposed to capture the hierarchical structure of graphs. In \citet{poincaremaps}, Poincaré embedding~\citep{poincare} has been applied to extract hierarchies from biological cell data.

However, there are two major problems with existing graph embedding methods that are used for extracting the hierarchical structures.

The first problem concerns the selection of the origin point. The distance from the origin is used as an indicator of hierarchy when applying the embedding method to the space where an isometric map exists. In non-Euclidean spaces, existing graph embedding methods use a loss function that depends on the distance between two nodes in the embedding.
Because the distance from the origin is changed by the isometric transformation as the value of the loss function is not changed, it is not appropriate to use the distance from the origin as an indicator of hierarchy. In other words, when we define the indicator of the hierarchy in the space where the isometric map exists, it is necessary to be invariant to the isometric map.

The second problem is a lack of extensibility.
If we need to solve a problem by adding hierarchical information to a graph embedding that has already been learned for another purpose, you will need to learn another embedding method or solve the problem as an independent process. Therefore, we cannot directly apply existing embedding method for extracting the hierarchical structure.


In this paper, we propose a method of embedding graphs in metric cones as a solution to the two problems described above.
A metric cone can be defined on any length metric space---a wide class of metric spaces (including normed vector spaces, manifolds, and metric graphs). This space's dimension is larger than that of the original space. It is known that the curvature of this space can be varied, and a method of changing the structure of the data space for analysis has also been proposed \citep{kei2019}.

First, we show that it is effective to use the coordinate corresponding to ``the height of the metric cone" (a one-dimensional parameter added to the original space) as an indicator of hierarchy.
Second, we demonstrate that the hierarchical structure can be extracted by learning the embedding of the metric cone, which optimizes the  only one-dimensional parameter, in case we have a graph embedding pretrained. By keeping the coordinates of the original space fixed, we show that only the one-dimensional parameter corresponding to the height of the cone is learned, and the computation time is reduced.

In cone embedding, we show that the curvature  of the embedding space can be optimized by data driven method , like \citet{wilson} and \citet{chami}. The curvature of a metric cone varies with one parameter that corresponds to the generatrix of the cone. If we have the pretrained embedding, we show that the learning method is also computationally efficient, because curvature can be optimized by learning one-dimensional parameters without changing the structure of the original space (i.e., without learning the embedding in the original space).

The remainder of this paper is organized as follows.
Section 2 describes related research. In Section 3, we propose the method of graph embedding in a metric cone, with the introduction of 1) graph embedding in non-Euclidean spaces and 2) the definition and properties of metric cones.
Section 4 presents the experimental results using graph data, followed by a conclusion and future perspectives in Section 5.

\section{Related Work}
\subsection{Random Walk Models}
DeepWalk~\citep{deepwalk} samples series by random walks on graphs and applies a method for embedding series data, such as the skip-gram model (word2vec~\cite{word2vec}). The sampling method of DeepWalk's random walk series has been modified to depend on edge weights in node2vec~\citep{node2vec}, and a method that simultaneously optimizes hyperparameters (such as random walk length) has been proposed in ~\cite{watchyourstep}.
Unlike our proposed method, these methods were not designed to represent a hierarchical structure but were simply designed to embed the structure of a graph with high accuracy, which is evaluated by predicting edge links.

\subsection{Dimensionality Reduction Models}
Dimensionality compression methods exist for data in Euclidean space,
such as multidimensional scaling (MDS)~\citep{mds},
IsoMAP~\citep{isomap}, and locally linear embedding (LLE)~\citep{lle}.
By applying it to the adjacency matrix of a graph, an embedded representation of the graph can be obtained.
There are also methods to obtain an embedded representation of a graph by applying dimensionality reduction to the graph's Laplacian matrix~\citep{le} instead of the adjacency matrix. As an alternative to random walk models, these graph embedding methods are also used to accurately embed the structure of the graph but not to represent a hierarchical structure.

\subsection{Graph Neural Networks}
Methods to compress dimensionality of adjacency matrices~\citep{dngr}\citep{sdne} or the graph's Laplacian matrices~\citep{gae}  using a neural network (autoencoder) to obtain a representation of the graph embedding have been proposed. 
These methods can be easily applied to other tasks. However, as with the above categories, these methods are designed for simply embedding the structure of a graph but not for representing a hierarchical structure.

\subsection{Non-Euclidean Models (Hyperbolic Embedding Models)}
Poincaré embedding~\citep{poincare} is a method to embed the adjacency matrix of a graph in a skip-gram model, large-scale information network embedding (LINE)~\citep{line}, which is constructed on a Poincaré sphere. In addition, there are other methods to embed graphs into Lorentz models~\citep{lorentz} and to embed each node of a graph as a cone instead of a point~\citep{pmlr-v80-ganea18a}.
These methods exist to describe the hierarchy of each graph according to the inclusion relation of the cones.
In previous research, the hierarchical structure could be accurately captured, but the problem of invariance for isometric maps prevented the natural definition of a metric that would represent the hierarchical structure. 
Therefore, embedding parameters in non-Euclidean spaces makes them difficult to use in other tasks. In contrast, our proposed method can use a Euclidean embedded representation of all parameters except for the one-dimensional parameters. The purpose is to provide an embedding method that is easy to use for other tasks and provides a natural indicator of hierarchical information.

\section{Methods}
From this point onwards, the set of edges in an undirected graph $G$ is denoted by $E$, the set of vertices by $V$, and the embedded space by $X$.
\subsection{Graph Embedding in Non-Euclidean Spaces}
Following Poincaré embedding, we learn the embedding of a graph $G$ by maximizing the following objective function: 
\begin{align}
    L = \sum_{(u,v)\in E}\log\frac{\exp\left(-d(u,v)\right)}{\sum_{v'\in N^c(u)}\exp\left(-d(u,v')\right)},
\end{align}
where $N^c(u):\{v'\in V| (u,v') \notin E\}$ denotes the set of points not adjacent to node $u$ (including $u$ itself), and $d$ denotes the distance function of the embedded space (for Poincar\'e embedding and the Poincaré sphere).
This objective function is a negative sampling approximation of a model in which the similarity is $-1$ times the distance, and the probability of the existence of each edge is represented by a SoftMax function on the similarity.
The maximization of the objective function is done by stochastic gradient descent on Riemannian manifolds (Riemannian SGD). The stochastic gradient descent over the Euclidean space updates the parameters as follows:
\begin{align}
    u \leftarrow u - \eta \nabla_u L(u),
\end{align}
where $\eta$ is the learning rate.
However, in non-Euclidean, the sum of vectors is not defined, and $\nabla_u L(u)$ is the point of the tangent space $T_u X$ of $u$; hence, SGD cannot be applied.
Therefore, we update the parameters by using an exponential map instead of the sum:
\begin{align}
    u \leftarrow \exp_u (- \eta \nabla_u^R L(u)).
\end{align}
With the metric tensor of the embedding space as $g_u(u\in V)$, the gradient on the Riemannian manifold $\nabla_u^R L(u)$ is the scaled gradient in the Euclidean space:
\begin{align}
    \nabla_u^R L(u) = g_u^{-1}\nabla_u L(u).
\end{align}

\subsection{Metric Cone}
The metric cone is similar to ordinary cones (e.g. circle cones) in the sense that it is defined as a collection of line segments connecting an apex point to a given set.
However, the metric cone has a notable property such that every point in the original set is embedded at an equal distance from the apex point and this is a desirable property for hierarchical structure extraction.

\begin{figure*}[htb]
    \centering
    \includegraphics[width=10cm]{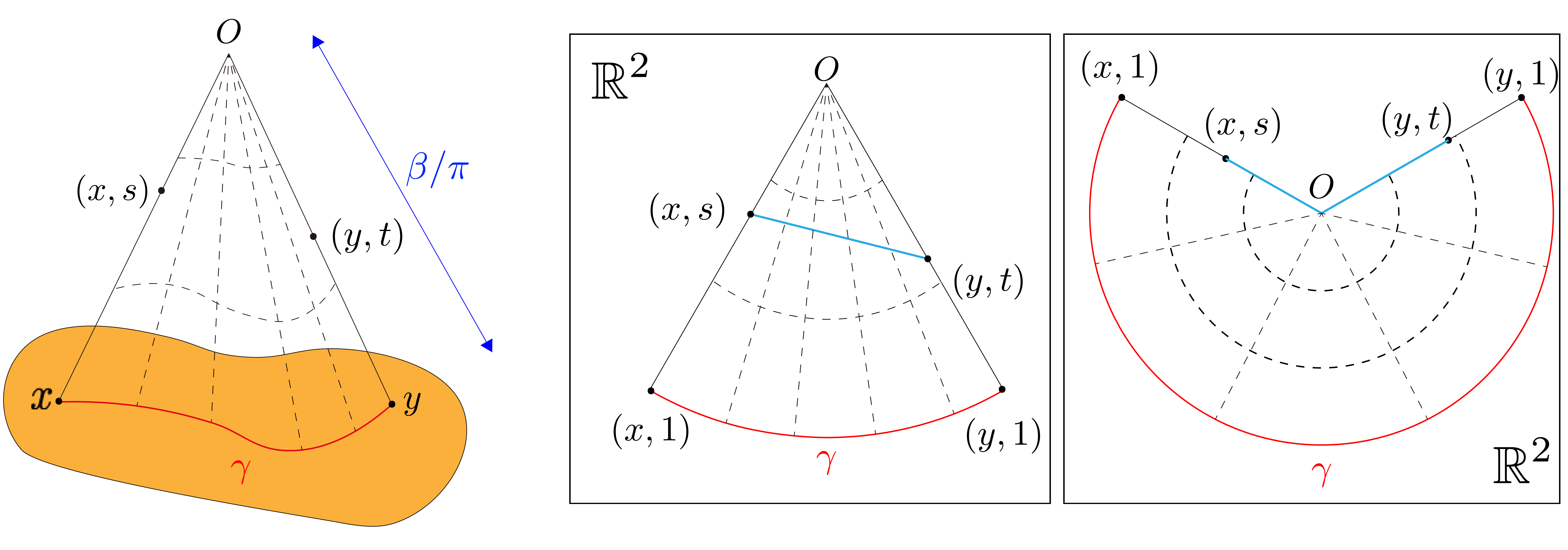}
    \caption{The Left figure depicts a conceptual image of an original space and its metric cone. A circle section to compute the distance in the metric cone is depicted in the middle figure (when the apex angle $< \pi$) and the right figure (when the apex angle $\geq \pi$).}
    \label{fig:metric_cone}
\end{figure*}

The metric cone has been studied as an analogy to the length metric spaces of the tangent cone for differential manifolds with singularities.
Length metric space is a metric space where the distance between any two points is equal to the shortest curve length connecting them. 
Length metric space includes Euclidean spaces, normed vector spaces, manifolds (e.g., Poincaré ball; sphere), metric graphs, and many other metric spaces.
Assume the original space $Z$ is a length metric space, then the metric cone generated by $Z$ is $X:= Z\times[0,1]~/~Z \times \{0\}$ with a distance function determined as follows:
\begin{align}
    \tilde{d}_{\beta}&((x, s),(y, t))\nonumber\\
    &=\beta \sqrt{t^{2}+s^{2}-2 t s \cos \left(\pi \min \left(d_{Z}(x, y) / \beta, 1\right)\right)}
\end{align}
where $\beta>0$ is a hyperparameter corresponding to the length of the conical generatrix. 
Note that the metric cone itself also becomes a length metric space, and it embeds the original space (i.e., the space is one dimension larger than the original space).
The distance in the metric cone corresponds to the length of the shortest curve on the circle section (blue line segment(s) in the right two subfigures in Figure \ref{fig:metric_cone})  whose bottom circumference is the distance of the original space $Z$ and whose radius is $\beta$. 

When the curvature is measured in the sense of CAT($k$) property, a curvature measure for general length metric spaces, the curvature value $k$ can be controlled by $\beta$. Other properties of the metric cone are examined in \citet{cat0}, \citet{deza}.
Because the metric cone can change the curvature of the space by changing parameter $\beta$, its usefulness has been reported in an analysis using the structure of the data space~\citep{kei2019}.

The metric $\tilde g$ of a metric cone is obtained by calculating the two-time derivative of the distance as follows (see supplementary materials for more details):
\begin{equation}
    \tilde{g}_{(x,s)}=\left(\begin{array}{cc}\pi^{2} s^{2} g_x & 0 \\ 0^\top & \beta^{2}\end{array}\right),\label{eq:cone_metric}
\end{equation}
where $g_x$ represents the metric of $Z$ at $x$. Combining this metric and the argument in  Sec3.1, the algorithm of cone embedding can be described as Algorithm 1.
\begin{figure}[!t]
\begin{algorithm}[H]
    \caption{Learn the cone embedding \{$(u,s)$\}}
    \label{alg1}
    \begin{algorithmic}[1]   
    \REQUIRE \text{graph} $G = (V, E)$, \text{cone's hyperparameter} $\beta$ \\
    and the pretrained embedding $\{x\}$  in original space $Z$
    \ENSURE {\text the cone embedding }\{$(u,s)$\}
    \STATE calculate the distance matrix $D=(d_{ij}),\ d_{ij}=d_Z(x_i,x_j)$\\
    \STATE minimize the loss function:\\
    (calculate efficiently by referencing the distance matrix $D$)
    \begin{align*}
    L =& \sum_{((u,s),(v,t))\in E}{-d((u,s),(v,t))}\\
    & -\log\left(\sum_{(v',t')\in N^c((u,s))}\exp\left(-d((u,s),(v',t'))\right)\right)
    \end{align*}
    via riemannian stochastic gradient descent:
    \begin{align*}
        (x,s) \leftarrow \operatorname{proj}((x,s) - \eta \tilde{g}^{-1}\nabla L)
    \end{align*}
    \end{algorithmic}
\end{algorithm}
\end{figure}
Instead of  the exponential map of metric cone, we use the first-order approximation using $\operatorname{proj}(x,s)$:
\begin{align}
   \operatorname{proj}(x,s) =\left\{ \begin{array}{ccc}
      (x,s)  & \text{if}& \epsilon<s<1-\epsilon, \\
      (x,1-\epsilon)    & \text{if}& s\geq 1-\epsilon ,\\
      (x, \epsilon) & \text{if}& s \leq \epsilon.
   \end{array}
   \right.
\end{align}

\subsubsection{Score Function of Hierarchy}
The Poincaré embedding defines an index in \citet{poincare}, which is aimed to be an indicator of the hierarchical structure, that depends on the distance from the origin:
\begin{align}
\label{eq:eq1}
    \operatorname{score(u, v)}=-\alpha(\|v\|-\|u\|) d(u, v)
\end{align}
This score function is penalized by the part after $\alpha$, so if $\mathbf v$ is closer to the origin than $\mathbf u$, then it is easier to obtain larger values. In other words, $\mathbf v$ is higher in the hierarchy than $\mathbf u$ (i.e., ``$\mathbf u$ is a $\mathbf v$'' relationship holds).
However, it is not appropriate to use this indicator for the Poincaré embedding. This model learns the embedding by maximizing (1)
, where 
\begin{align}
d(x,y):=\operatorname{arcosh}\left(1+\frac{\|x-y\|^2}{(1-\|x\|^2)(1-\|y\|^2)}\right).
\end{align}
This loss function only depends on the distance between the two embeddings.
However, an isometric transformation in Poincar\'e ball exists, known as Möbius transformation \citep{mobius}. Möbius transformation is defined as a map $f:\mathbb B^n \text{(open unit ball)} \rightarrow \mathbb B^n$, which can be written as a product of the inversions of $\bar{\mathbb R}^n(:=\mathbb R^n \cup \{\infty\})$ through a sphere $S$ that preserves $\mathbb  B^n$.

In contrast to Poincar\'e ball, the isometric transformation on the metric cone does not exist when the coordinate in the original space is fixed (we prove this property in Section 3.2.2). 
When we embed a graph into a metric cone, we define an indicator of the hierarchical structure by replacing the norm with a coordinate corresponding to the height of the cone (a one-dimensional parameter added to the original space):
\begin{align}
    \operatorname{score((\mathbf u, s), (\mathbf v, t))}=-\alpha(s - t) d(\mathbf u, \mathbf v).
\end{align}
A point closer to the top of the cone is higher in the hierarchy, which is natural for representing hierarchical structure.
\subsubsection{Identifiability of the Heights in Cone Embedding}
As mentioned above, for Poincar\'e embedding, there is an isometric transformation on the Poincar\'e ball, and the heights of the learned hierarchy are not invariant to such transformation. Here, we show that such a phenomenon does not occur for the cone embedding, i.e., the heights of the hierarchy are (almost) uniquely determined from the distance between the embedded data points in a metric cone.

Let $Z$ be an original embedding space (connected length metric space), and let $X$ be a metric cone of $Z$ with a parameter $\beta>0$.
We assume that each data point $z_i \in Z~(i=1,\dots,n)$ has its specific ``height'' $t_i \in [0,1]$ in the metric cone $X$.
Our proposed method embeds data points into a metric cone based on the estimated distances $\tilde{d}_\beta(x_i,x_j)~(i,j=1,\dots,n)$ and tries to compute the heights $t_1,\dots,t_n$ as a measure of  the hierarchy level.
However, it is not evident if these heights are identifiable only from the information of the original data points in $Z$ and the distances $\tilde{d}_\beta(x_i,x_j)~(i,j=1,\dots,n)$ in the metric cone.
The following theorem guarantees some identifiability.
\begin{theorem} \label{thm:1}
\begin{enumerate}
\setlength{\itemsep}{-0.1cm}
\item[(a)]
Let $n\geq 3$ and assume that $z_1,\dots, z_n$ are not all aligned on a geodesic in $Z$. 
Then, the heights $t_1,\dots,t_n$ are identifiable up to at most four candidates.
\item[(b)]
Let $n\geq 4$ and assume $z_1,\dots,z_n$ and $t_1,\dots,t_n$ take ``general'' positions and heights, respectively.
Then, the heights $t_1,\dots,t_n$ are identifiable uniquely.
\item[(c)]
If $d_Z(z_i,z_j)\geq \frac{\beta}{2}$ for all $i,j=1,\dots, n,~i\neq  j$, 
then the heights  $t_1,\dots,t_n$ are identifiable uniquely.
\end{enumerate}
\end{theorem}
A rigorous version of Theorem \ref{thm:1}, including
the precise meaning of ``general'' in (b), is explained in supplementary materials.
Theorem \ref{thm:1}(a) indicates that the candidates of heights are finite, and we can expect the algorithm to converge to one of them, except for a very special data distribution in the original space $Z$.
Moreover, by (b), even the uniqueness can be proved under very mild conditions.
The statement in (c) implies that the uniqueness holds for arbitrary data distributions when we set $\beta$ sufficiently small.

Remark that the assumption of  ``general''  positions in Theorem \ref{thm:1} (b)
is satisfied easily for most data distributions.
For example, if both $z_1,\dots,z_n\in \mathbb{R}^d$ and $t_1,\dots,t_n\in [0,1]$ are i.i.d. from a probability distribution whose density function exists with respect to the Lebesgue measure, then it is easy to see the assumption holds almost surely and
therefore uniqueness of the solution is guaranteed.
Note that for $n=3$ under the same setting,  there can be multiple solutions with a positive probability.

\subsubsection{Using Pretrained Model for Computational Efficiency and adaptivity for adding hierarchcal information}
Consider a situation where we already have a trained graph embedding on a Euclidean space (e.g. LINE \citep{line}), and we try to learn the embedding in a metric cone of the Euclidean space to extract information about the hierarchical structure.
In this case, we can reduce the computational cost by fixing the coordinates corresponding to the original Euclidean space and learn only the one-dimensional parameters corresponding to heights in a metric cone added to the original space because the metric cone is one dimension larger than the original space. The distance between each embedding in the original space is calculated beforehand, since no updates are made by learning except for the 1D parameter to be added. By referring to the pre-computed distances in the original space when calculating the distances between each embedding on the metric cone ($d_z(x, y)$ in Eq. (6)) during training, we can reduce the amount of computation for the parts other than preprocessing, regardless of the dimension. In addition, because the embedding in the original Euclidean space is preserved, it can be used as an input to the neural network (when the task considers information about the hierarchy, and the added one-dimensional parameters are also used as input) and can be easily applied to other tasks. However, other non-Euclidean embedding methods to extract hierarchical structures are not scalable because these methods cannot be applied directly to solve other tasks. For example, deep neural networks cannot use a non-Euclidean embedding as input because the sum of two vectors in the space and scalar product is not generally defined.
\subsubsection{Comparison with hierarchical clustering}
It has been shown by \citet{sarkar} that the embedding of tree-structured (undirected) graph data can be done naturally in hyperbolic space, but graph data with hierarchical structure does not necessarily have a tree structure in general. However, graph data with a hierarchical structure does not have a tree structure in general(e.g. there can be a cycle when a child node has two parents which have the same parent.), so the combination with  hyperbolic embedding and hierarchical clustering may not be suitable in such cases. Cone embedding does not assume the tree structure and extracts the hierarchical structure by using the property that the closer to the origin $O$, the shorter the distance of between data points,so that embedding can be learned even in this situation.
\subsubsection{Variable Curvature}
One of the essences of Poincar\'e embedding is that a negative curvature of the Poincar\'e sphere 
is suitable for embedding tree graphs.
The curvature of a metric cone has a similar property, i.e. a metric cone has a more negative curvature than the original space and, furthermore, the curvature can be controlled by hyperparameter $\beta$.
We will verify these facts mathematically from two different aspects:
(i) the scalar and the Ricci curvatures of a Riemannian manifold and (ii) the CAT($k$) property of a length metric space.

First assume the original space $\mathcal{M}$ is a $n$-dimensional Riemannian manifold with a metric $g$. 
Then the metric $\tilde{g}$ of the corresponding metric cone with $\beta$ can be defined
except the apex and it becomes as (\ref{eq:cone_metric}). 
Let $1,\dots,n$ be coordinate indices corresponding to $x\in \mathcal{M}$ and $0$ be the index corresponding to $s\in [0,1]$.
The Ricci curvatures$\tilde{R}_{ij}$ and the scalar curvature $\tilde{R}$ at $(x,s)$ become
\begin{align}
\tilde{R}_{\alpha\gamma}&=R_{\alpha\gamma}-\pi^2 (n-1) \beta^{-2} \tilde{g}_{\alpha\gamma},\\
\tilde{R}_{\alpha0}&=\tilde{R}_{0\alpha}=\tilde{R}_{00}=0, \label{eq:ricci_curvature}\\
\tilde{R}&=\{\pi^{-2}R-n(n-1)\beta^{-2}\}s^{-2} \label{eq:scalar_curvature}
\end{align}
where $\alpha,\gamma$ are coordinate indices in $1,\dots,n$ and $R_{ij}$ and $R$ are the Ricci curvatures and the scalar curvature of $\mathcal{M}$, respectively.
See the supplementary materials for the derivation of such curvatures.
The scalar curvature and the Ricci curvatures $\tilde{R}_{\alpha\gamma}$ becomes more negative than (a constant times of) the original curvature for $\beta<\infty$ and $n\geq 2$. 
Moreover, the smaller value of $\beta$ makes the curvature more negative thus it becomes possible to control the curvature by tuning $\beta$.
Note that the closer to the apex, i.e. the smaller the value of $s$, the greater the change of the scalar curvature.

Second assume the original space $\mathcal{M}$ is a length metric space.
This doesn't require a differentiable structure and is more general than the Riemannian manifold.
In this case, we cannot argue the curvatures using the Riemannian metric but
the CAT($k$) property can be used instead.
In \citet{kei2019}, they proved the curvature of the metric cone is more negative or equal to the curvature of the original space and it can be controlled by $\beta$ in the sense of the CAT($k$) property.
\section{Experiments}
The claim in this paper is that “a hierarchical structure can be captured by adding a one-dimensional parameter and
embedding it in a metric cone.” Therefore, we evaluate the proposed method in two experiments:
\begin{itemize}
    \item prediction of edge direction for artificial directed graphs
    \item estimation of hierarchical score by humans for WordNet.
\end{itemize}
As a comparison, we compare the proposed method with three other methods: Poincaré embedding\citep{poincare}, 
and ordinary embedding in Euclidean space, which are known to capture the hierarchical structure of graphs.
For the Euclidean embedding, we use the distance from the mean of embedded data points as the hierarchical score in (8).

\subsection{prediction of edge direction for directed graphs}
In this experiment, we estimate the orientation of directed edges for some simple graphs.
We use the following three patterns of graphs with a naturally set hierarchical structure:

\begin{itemize}
    \item Graphs generated by a growing random network model called Barabási–Albert preferential attachment\citep{bamodels}  with $m=2$, where $m$ is the number of edges to attach from a new node to existing nodes
    \item Complete $k$-ary tree
    \item Concatenated tree of two complete $k$-ary sub-trees
\end{itemize}
For each tree, node depth can be treated as its hierarchy.
The concatenated tree is created by connecting the roots of two complete $k$-ary trees to a new node, which is then used as the new root.
The concatenated tree is considered to study the effect of node degree on the cone embedding as will be explained below.
\begin{figure*}[htb]
    \begin{tabular}{ccc}
\begin{minipage}[t]{0.3\hsize}
        \centering
    \includegraphics[width=5cm, height=5cm, ]{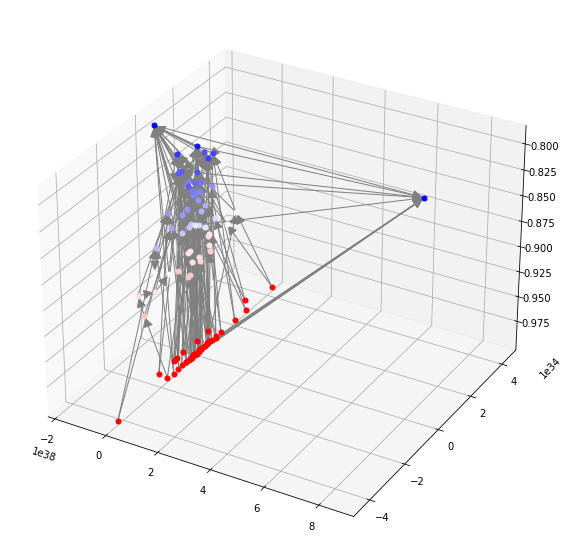}
     \end{minipage}
     &  
     \begin{minipage}[t]{0.3\hsize}
     \centering
    \includegraphics[width=5cm, height=5cm, ]{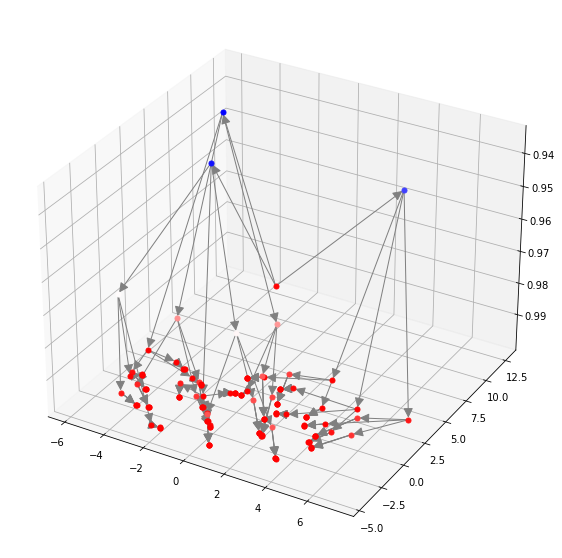}
          \end{minipage}
          &  
     \begin{minipage}[t]{0.3\hsize}
     \centering
    \includegraphics[width=5cm, height=5cm, ]{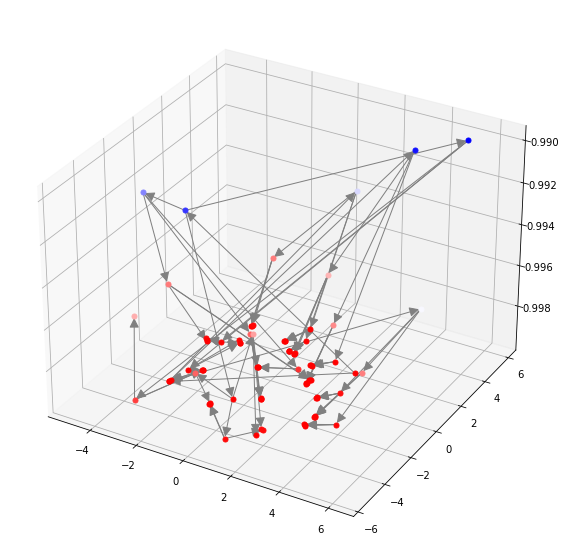}
          \end{minipage}
     
\end{tabular}
\caption{Graphs used for training: (left) model trained by Barabási–Albert model, (middle) complete $k$-ary tree, (right) concatenated tree of two complete $k$-ary trees.
The $x$- and $y$-axes represent embedding in Euclidean space, which is dimensionally reduced to two dimensions by principal component analysis, and the $z$-axis represents height in metric cone (coordinates representing hierarchy).}
\label{fig:data_graphs}
\end{figure*}

In this experiment, we learn the embedding of each undirected graph.
For each undirected edge of the learned graph, we estimate the direction by the computed hierarchical scores $score(u,v)$:
\begin{align}
  \text{total\_score}&=\sum_{(u,v)\in E} \tilde{score}(u,v)/|E| \\
  \tilde{score}(hypo,hype) &= \left\{\begin{array}{cc}
        1 &  score(hypo,hype) > 0\\
        0 &  \mathrm{otherwise}
    \end{array}\right.\\
    \text{where } hype: &\text{ higher hierarchy node}\nonumber\\
   hypo:&\text{ lower hierarchy node}\nonumber
\end{align}

The experimental results are shown in Table \ref{tab:euclid} and the cone embedding shows overall good and stable estimation accuracy for hierarchies. 
Examples in Figure \ref{fig:data_graphs} depict how each graph is embedded in a metric cone.
The reason for the instability in the accuracy of Poincaré embedding may be the lack of invariance with respect to the equidistance transformation, as we have explained. 
The main reason for the poor hierarchical estimation accuracy of Euclidean embedding is that the root or higher hierarchical nodes are embedded apart from the cluster of other nodes as in Figure \ref{fig:emb_tree}.
As a result, the root becomes far from the mean of the embedded points.

\begin{figure}
    \centering
    \includegraphics[width=6cm]{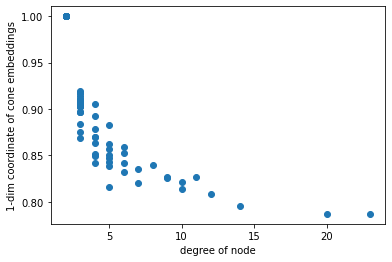}
    \caption{Plot of the hierarchy value of each node in a cone embedding (a newly added one-dimensional parameter) against their node degree.}
    \label{fig:degree}
\end{figure}
For the Barabási–Albert model, the relationship between the (added 1D) coordinates of the cone hierarchy and the order is shown in Figure \ref{fig:degree}.
We can see that there is a strong relationship between the degree and the hierarchy.
This raises the suspicion that the degree of the node alone determines the hierarchy of the embedding. 
However, the fact that cone embedding provides high estimation accuracy even for the concatenated trees with low root degree indicates that this is not true.

\begin{figure}
    \centering
    \includegraphics[width=6cm]{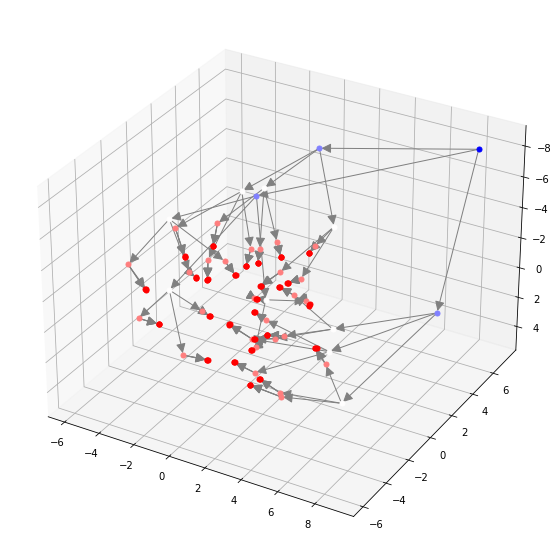}
    \caption{An embedding of the complete $k$-ary tree ($k=3$).\\
    Each point is plotted by the 3D Euclidean embedding, and the color represents length of shortest path from root node (the bluer the color, the higher the hierarchy).}
    \label{fig:emb_tree}
\end{figure}

\begin{table*}[hbt]
     \centering
      \caption{Result for prediction of edge direction for directed graphs. (we list accuracy and standard deviation)
      }
     \begin{tabular}{|c||c|c|c|c|c|}
     \hline 
     Model &Barabási–Albert &\multicolumn{2}{|c|}{complete k-ary-tree}&\multicolumn{2}{|c|}{Concatenated tree of two k-ary trees}
      \\
      & (100nodes) &\multicolumn{1}{c}{k=3(121nodes)} &\multicolumn{1}{c|}{k=5(781nodes)} & \multicolumn{1}{c}{k=3(81nodes)} & \multicolumn{1}{c|}{k=5(313nodes)}\\\hline
  Metric Cone    &0.936(se:0.005) &0.787(0.049) & 0.799(0.037) & 0.783(0.045)&0.744( 0.056) \\\hline
   Euclidean      & 0.181(0.004) &0.074(0.088) &0.127(0.020)  &0.190(0.136) & 0.155(0.031)\\\hline
    Poincar\'e &0.957(0.012) & 0.935(0.015) &0.351(0.022)  &0.880(0.060) & 0.606(0.043) \\\hline
     \end{tabular}
     \label{tab:euclid}
\end{table*}
\begin{table*}[htb]
    \centering
    \caption{MAP, Mean Rank score, Hyperlex score(correlation efficient) and computation time for the Reconstruction and Link Prediction  setting in WordNet. Since the link prediction settings is to estimate generalization performance, the computation time and correlation with human score are omitted.}
    \small
    \begin{tabular}{|c|c||c|c|c|c||c|c|c|c|}
    \hline 
    Model &evaluation &\multicolumn{4}{|c|}{Reconstruction}&\multicolumn{4}{|c|}{Link Prediction}
     \\\cline{3-10}
     & metric &10 & 20 & 50 & 100& 10 & 20 & 50 & 100\\\hline
    Euclidean  & MR& 3.16 &3.16 & 3.16&3.16 & 29.37&8.18 &8.16 &8.16\\\cline{3-10}
     & MAP &0.999 &0.999 & 0.999&0.999 &0.745 &0.990&0.996 &0.997 \\\cline{3-10}
      &corr &0.259 &0.423 &0.464 & 0.478&- &- &- &-\\\cline{3-10}
         &comp. time &9958.1 &12126.0 &12062.2 & 12043.5&- &- &- &- \\\hline
    Poincar\'e  & MR& 27.77& 26.62&25.74&26.12 &415.34 & 389.34&382.65 &380.41\\\cline{3-10}
     & MAP &0.879 &0.881&0.883& 0.882 & 0.721 & 0.727&0.729&0.729 \\\cline{3-10}
      &corr &  0.297&0.299 &0.290 &0.285 &- &- &- &-\\\cline{3-10}
         &comp. time &12250.1 &12152.4 &12180.1 &20795.6 &- &- &- &- \\\hline
     
    Cone & MR& 3.16 &3.16 &3.16 &3.16 & 27.85 & 8.18 & 8.16&8.16\\\cline{3-10}
    (Our Model)& MAP &0.999 &0.999 & 0.999&0.999 &0.745 &0.990&0.996 &0.997 \\\cline{3-10}
      &corr & 0.379&0.470 &0.483 &0.490 &- &- &- &-\\\cline{3-10}
         &comp. time &10901.2 &11289.6 &11431.0 &12164.5 & -&- & -&- \\\hline
    \end{tabular}
    \label{tab:graph_emb3}
\end{table*}

\subsection{Embedding Taxonomies}
Following an experiment in \citet{poincare} for the Poincaré embedding, we evaluate the embedding accuracy of the hierarchical structure using WordNet.
To verify this, we embed the nouns in WordNet into a metric cone and use the score function (10).
Note that hyperparameter $\alpha$ was set to $10$.
The output of this score function and the correlation coefficient of the HyperLex dataset's score (evaluated manually whether a word is a subword of another word) are used to evaluate the ability to represent the hierarchical structure of the model.
Also we evaluate the effectiveness of proposed method by two setups, reconstruction and link prediction.
In the reconstruction setup, all of the graph data are used for training, and the results are evaluated according to the accuracy with which the graph is reconstructed from the learned embedding. Because the same data are used for training and evaluation, we evaluate the fittingness of the embedding method to the data (not generalization performance).
In the link prediction setup, edges are randomly removed for training. In the evaluation, we use the same criteria as in the reconstruction setting to evaluate the accuracy of the removed edges.
The accuracy is calculated as follows:
\begin{enumerate}
    \item Fix one node and calculate the distance to all other nodes. Sort the nodes in order of proximity.
    \item We sort all nodes except the reference node in order of distance proximity, and calculate the average value and average precision of the rankings of neighboring nodes.
    \item Perform the above two operations on all the nodes and take the average.
\end{enumerate}
since their experiments reported better accuracy when euclidean embedding is trained without normalization to the unit vector after updating embedding by Riemannian SGD, we believe that this is due to this effect. But since there is a hierarchy in the graph to be embedded, only the edges connected to nodes that are neither root nor leaf are split into test data in the link prediction setting.
The embedding accuracy (mean rank (MR) and mean average precision (MAP)) and correlation coefficients are also shown in Table \ref{tab:graph_emb3}.
The table shows that our proposed model improves the score and captures the hierarchical structure better than other embedding methods.
Also, the time taken to learn the embedding is shown in the same table. The time units are in CPU time, and all the results are from the reconstruction settings and link prediction experiment. From the table, we can see that our method is efficient in learning and does not vary with the dimension of the embedding. This is due to the fact that the distance matrix is calculated beforehand, which avoids repeating useless calculations.

Furthermore, an example visualization of the hierarchical structure of the embedding vectors obtained by the training is shown in Figure 5. As the figure illustrates, the closer the coordinate corresponding to the height in the cone is to zero (closer to the top of the cone), the higher the noun in the hierarchy is located in the embedded representation. For visualization, the embedding vectors in Euclidean space were reduced to two dimensions by principal component analysis.

\begin{figure}[hbt]
        \centering
    \includegraphics[width=6cm, height=8cm, angle=90]{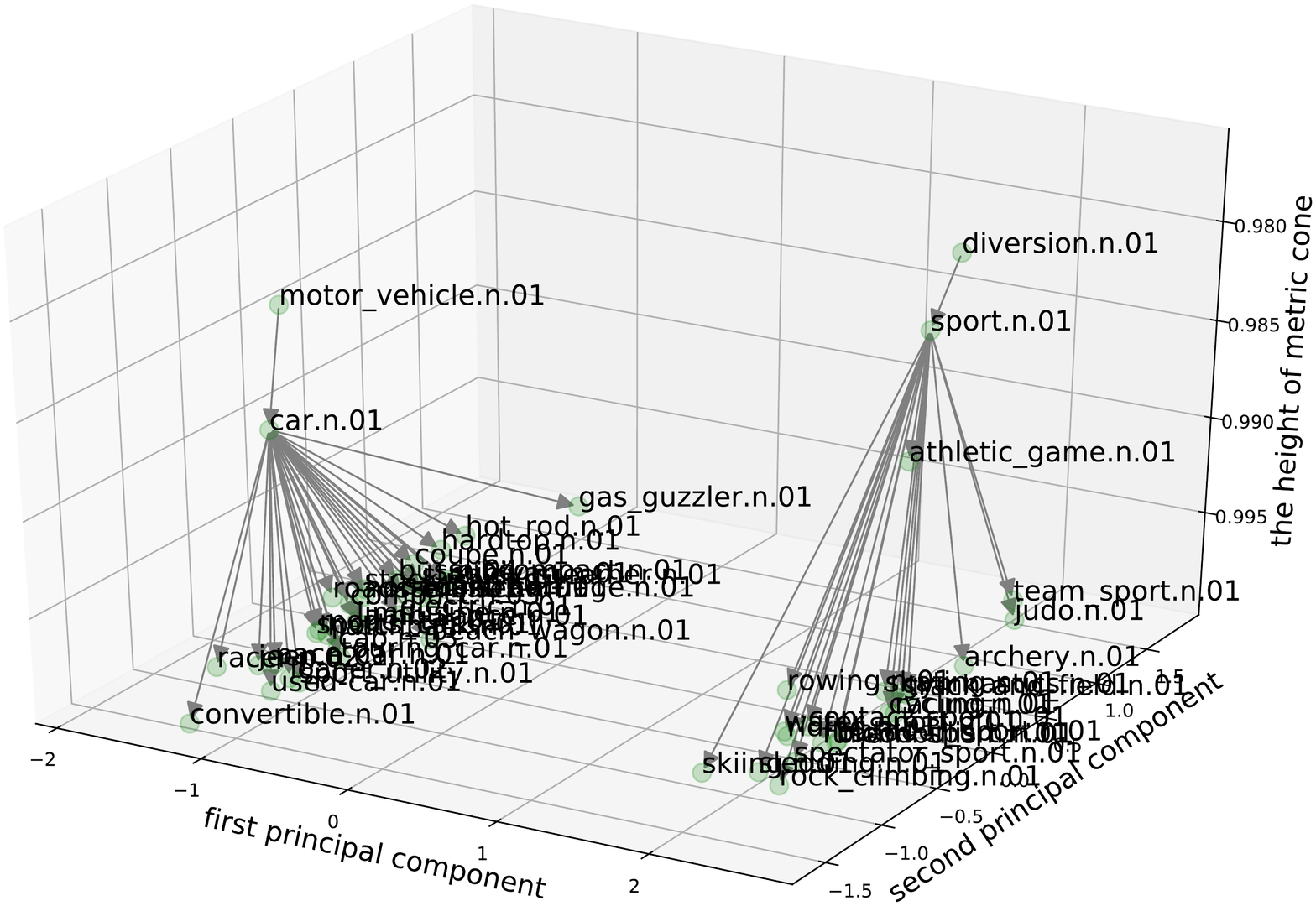}
    
\label{fig:my_label}
\caption{Visualization of WordNet Embedding using metric cone: each point is a word, and the two points connected by directed edges indicate that the word at the end of the arrow is a hyponym of the word at the start. }
\end{figure}

\section{discussion and future works}
In this study, we have demonstrated that a graph embedding in a metric cone that is a dimension larger than the existing embedding methods has the following advantages: 1) we naturally define an index (score function) as an indicator of hierarchy, 2) the proposed method has some adoptivity since it can introduce the hierarchy into various pretrained models by learning only newly added 1D parameters, and 3) thus, the optimization is computationally inexpensive.
By optimizing the 1D parameters, we have shown that the proposed method also has the flexibility to optimize the curvature to enhance the accuracy as well as other methods.
Future research topics include 1) efficient optimization of curvature, 2) development of an embedding method to update existing embeddings in learning, and 3) discovery of applications to other tasks.
In this study, we measured the improvement of the accuracy of curvature optimization by the grid search. However, more efficient methods, such as gradient-based methods, can be used to optimize the embedding space.
We also learned embedding in a metric cone under the constraint of not updating existing embeddings. 
This constraint can cause the optimization to fall into a local solution. 
Therefore, to optimize the entire embedding, an efficient method of optimizing functions on a metric cone (or Riemannian manifold) should be developed in future work.

\bibliography{main}

\appendix

\section{Derivation of the metric tensor of a metric cone}
Let $\mathcal{M}$ be an $n$-dimensional Riemannian manifold with a metric $g$. 
Then the metric $\bar g$ of the corresponding metric cone $\tilde{\mathcal{M}}=\tilde{\mathcal{M}}_\beta$ can be defined
except the apex.
Denote the square of the infinitesimal  distance in $\tilde{\mathcal{M}}$ as $|d\tilde{s}|^2$, then
\begin{align}
|d\tilde{s}|^2 
&= \bar d_\beta((x,r),(x+dx,r+dr))^2 \nonumber\\
&=\beta^2(2r^2 + 2rdr + dr^2 - 2(r + dr)\nonumber\\
 &\ \ \ r\cos(\pi \min (d_{\mathcal{M}}(x,x+dx)/\beta, 1))) \nonumber\\
&\approx \beta^2(2r^2 + 2rdr + dr^2 - 2(r^2+ r dr)\nonumber\\
&\ \ \ \left(1 - \frac{(\pi d_{\mathcal{M}}(x, x+ dx)/\beta)^2}{2}\right)) \nonumber\\
&\approx \beta^2dr^2 + {\pi^2r^2}\sum_{i,j}g_{ij}dx_idx_j \nonumber\\
& + \pi^2r dr \sum_{i,j}g_{ij}dx_idx_j \nonumber\\
&\approx \left(
\begin{array}{cc}
dx \\
 dr\end{array}
 \right)^\top \left(\begin{array}{cc}
(\pi^2r^2g_{ij}) & 0 \\
0 & \beta^2
\end{array}
\right)
\left(
\begin{array}{c}
  dx\\ 
  dr  
\end{array}
\right).
\end{align}
Therefore, the metric tensor $\bar g$ becomes
\begin{align}
\label{eq:mc_a1}
\bar g = \left(\begin{array}{cc}
r^2\pi^2g & 0\\
0 & \beta^2
\end{array}
\right).
\end{align}

\section{Derivation of the Ricci and the scalar curvatures of a metric cone}

We will derive the Ricci and scalar curvatures of metric cone $\tilde{\mathcal{M}}$
Let $0,1,\dots,n$ be the coordinate indices of metric cone $\tilde{\mathcal{M}}$ where 
$0$ corresponds to the radial coordinate $s\in (0,1)$ and $1,\dots,n$ correspond to $x\in \mathcal{M}$.

\begin{claim}\label{claim:1}
The Ricci curvatures $\tilde{R}_{ij}$ and the scalar curvature $\tilde{R}$ become
$$\tilde{R}_{\alpha\gamma}=R_{\alpha\gamma}-\pi^2 (n-1) \beta^{-2} g_{\alpha\gamma}, 
\tilde{R}_{\alpha0}=\tilde{R}_{0\alpha}=\tilde{R}_{00}=0,$$
\begin{equation}\tilde{R}=\{\pi^{-2}R-n(n-1)\beta^{-2}\}s^{-2}
\end{equation}
where $\alpha$ and $\gamma$ are coordinate indices in $1,\dots,n$ and $R_{ij}$ and $R$ are the Ricci curvatures and the scalar curvature of $\mathcal{M}$, respectively.
\end{claim}

\Proof
By Example 4.6 of \cite{janson2015riemannian},
if the metric of $\tilde{\mathcal{M}}$ is defined by the squared infinitesimal distance $|ds|^2$ in $\mathcal{M}$ and a $\mathcal{C}^2$-class function $w$ on an open interval $J\subset \mathbb{R}$
as
\begin{equation}
|d\tilde{s}|^2=\beta^2|dr|^2 + w(r)^2|ds|^2, \label{eq:ds2}
\end{equation}
the Ricci curvature tensor becomes
\begin{align}
\tilde{R}_{\alpha\gamma} 
= R_{\alpha\gamma} - \left((n-1)\left(\frac{w'}{w}\right)^2+\frac{w''}{w}\right)\tilde{g}_{\alpha\gamma}\nonumber\\
=R_{\alpha\gamma} - \left((n-1)\left(\frac{w'}{w}\right)^2+\frac{w''}{w}\right)w^2 g_{\alpha\gamma},
\end{align}
\begin{equation}
\tilde{R}_{\alpha 0} = 0,~~~
\tilde{R}_{0 0} = -(n-1)\frac{w''}{w}
\end{equation}

and the scalar curvature becomes
\begin{equation}
\tilde{R} = w^{-2}(R-n(n-1)(w')^2-2n w w'').
\end{equation}

Since the metric of a metric cone $\tilde{M}$ is given by
\begin{equation}
|d\tilde{s}|^2 = \beta^2 |dr|^2+ \pi^2 r^2 |ds|^2,
\end{equation}
by setting $\tilde{r}:=\beta r$ and $w(\tilde{r}):=\pi\beta^{-1}\tilde{r}$, we obtain the following form similar to (\ref{eq:ds2}):
\begin{equation}
|d\tilde{s}|^2=|d\tilde{r}|^2 + w(\tilde{r})^2|ds|^2. 
\end{equation}
By substituting $w(\tilde{r})=\pi\beta^{-1}\tilde{r}=\pi r$, $w'(\tilde{r})=\pi\beta^{-1}$ and $w''(\tilde{r})=0$, we obtain the Ricci and scalar curvatures in Claim \ref{claim:1}
\qed

\section{Identifiability of the hights in the cone embedding}
In this section, we will prove Theorem 1 of the main article.
Let us begin by rewriting Theorem 1 as a longer but more theoretically rigorous form.

\begin{theorem} [A rigorous restatement of Theorem 1]
\label{thm:a1}
Let $Z$ be a length metric space and $X$ be a metric cone of $Z$ with a parameter $\beta>0$.
Let $n$ be an integer at least 3. Fix $z_i \in Z$ and $x_i:=(z_i,t_i) \in X$ with $t_i \in [0,1]$ for $i=1,\dots,n$.
Denote a matrix $\tilde{D}:=[\tilde{d}_\beta(x_i,x_j)]_{i,j=1}^n$.
\begin{enumerate}
\item[(a)]
Assume $z_1,\dots, z_n$ are not all aligned in a geodesic. 
Given $z_1,\dots,z_n$ and $\tilde{D}$, 
the number of possible values of $(t_1,\dots,t_n)$ is at most four.
\item[(b)]
Let $n\geq 4$ and assume  $z_1,\dots,z_n$ and $t_1,\dots,t_n$ are in a ``general'' position.
Here ``general'' position means that, besides the assumption in (a), given any distinct 4 points $z_i, z_j, z_k, z_l\in Z$ and corresponding heights $t_i, t_j, t_k \in [0,1]$, but still $t_l$ can take infinitely many values.
Then $t_1,\dots,t_n$ are determined uniquely by $z_1,\dots,z_n$ and $\tilde{D}$.
\item[(c)]
If $d(z_i,z_j)\geq \beta/2$ for all $i,j=1,\dots, n,~i\neq  j$, 
then $t_1,\dots,t_n$ are determined uniquely by $z_1,\dots,z_n$ and $\tilde{D}$.
\end{enumerate}
\end{theorem}

Before the proof, we will state some remarks.

\begin{figure}[h]
\begin{center}
\includegraphics[height=4cm]{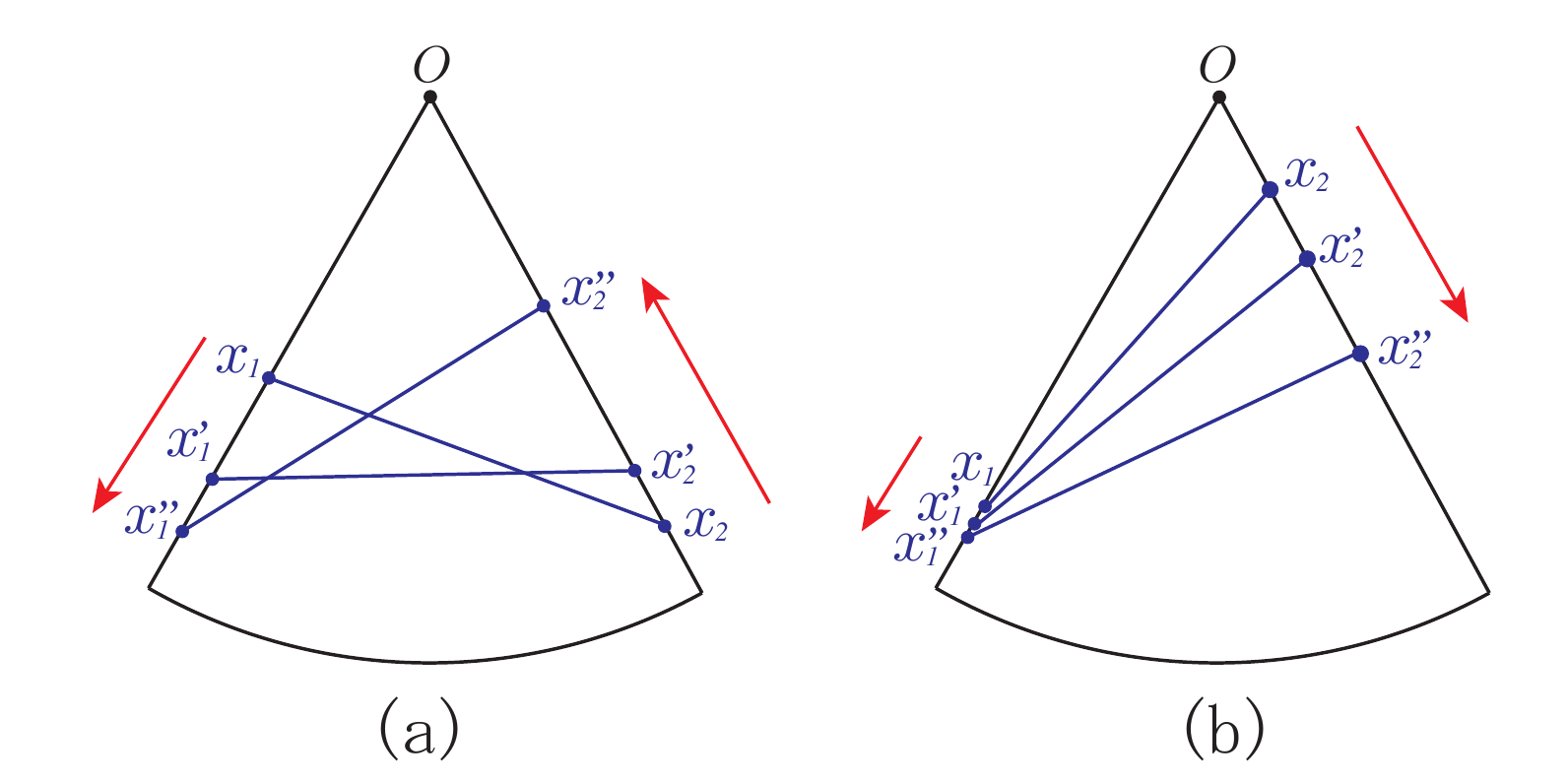}
\caption{Two types of movement for a line segment of constant length}
\label{fig:a1}
\end{center}
\end{figure}

If $n=2$, the identifiability problem reduces to an elementary geometric question:
given a circle sector as the right two subfigures of Figure 1 of the main paper and the length of the blue line segment(s) connecting $(x,s)$ and $(y,t)$, can $s$ and $t$ be determined uniquely?
The answer is evidently no.
But it is notable that there are two types of counterexamples.
The first type is as Figure \ref{fig:a1}(A), one point moves ``up'' and the other moves ``down''.
The other type as Figure \ref{fig:a1}(B) is maybe counter-intuitive: both moves ``up'' or ``down''.
Note that the second case does not happen if the angle $\theta$ is larger than or equal to $\pi/2$.

\begin{figure}
\begin{center}
\includegraphics[height=5cm]{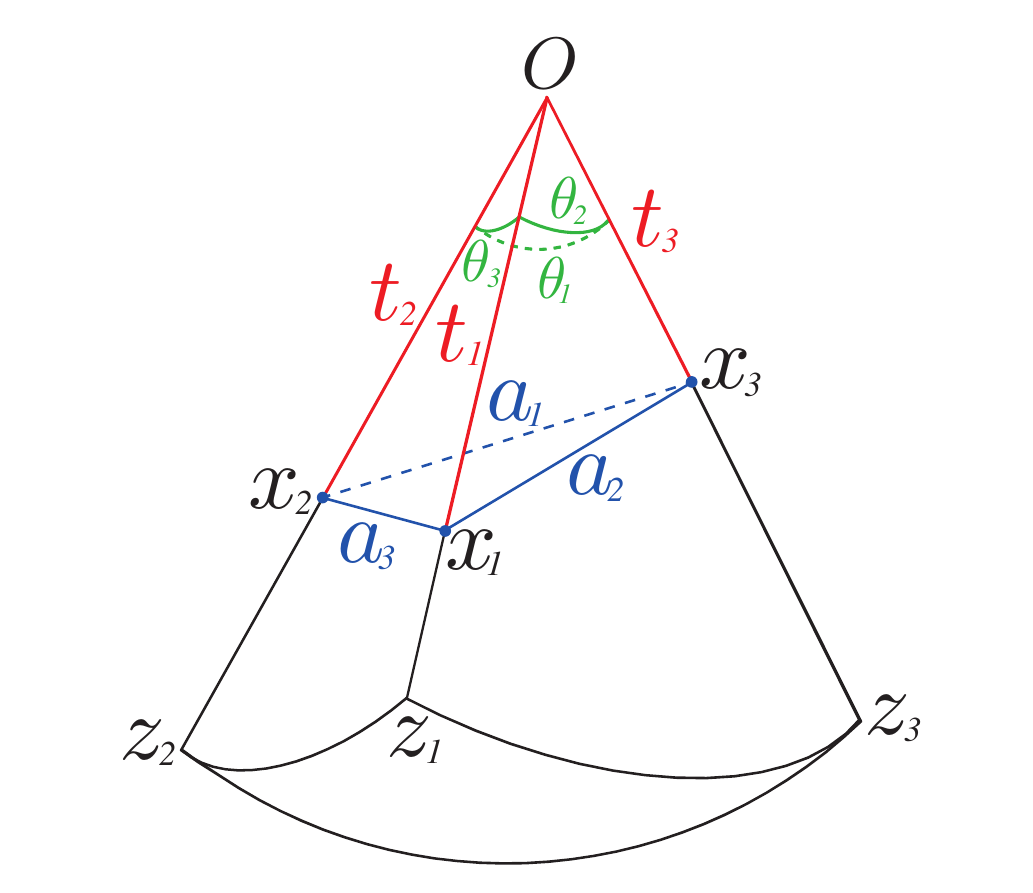}
\caption{Metric cone generated by three points $z_1,z_2,z_3\in Z$}
\label{fig:a2}
\end{center}
\end{figure}

If $n=3$, the picture becomes a tetrahedron as in Figure \ref{fig:a2}.
Here the angles and edge lengths are defined by
\begin{align} 
\theta_1&:=\pi \min(d_Z(z_2,z_3)/\beta,1), ~~a_1:=\tilde{d}_\beta(x_2,x_3) \nonumber \\ 
\theta_2&:=\pi \min(d_Z(z_3,z_1)/\beta,1), ~~a_2:=\tilde{d}_\beta(x_3,x_1)\label{eq:angles} \\ 
\theta_3&:=\pi \min(d_Z(z_1,z_2)/\beta,1), ~~a_3:=\tilde{d}_\beta(x_1,x_2)\nonumber 
\end{align}
and $\theta_1+\theta_2+\theta_3$ is assumed to be at most $2\pi$.
Then the geometrical question becomes
``when angles $\alpha,\beta, \gamma$ and edge lengths $a_1,a_2,a_3$ of  triangle $\triangle x_1 x_2 x_3$ is given, can the position of the points $x_1,x_2$ and $x_3$ be determined uniquely?''
If it is not unique and there are two different positions of $x_1,x_2$ and $x_3$,
at least one edge should move as in Figure \ref{fig:a1}(B) since it is impossible to move all the three edges as in Figure \ref{fig:a1}(A).
But if all of the angles are larger than or equal to $\pi/2$, this cannot happen.
This gives actually a geometrical proof of Theorem \ref{thm:a1} (c).

If $\theta_1+\theta_2+\theta_3$ is larger than $2\pi$, 
the geometric arguments become complicated.
We do not need this kind of case analysis when we use algebraic arguments as in the following proof.

Now we will prove the theorem.
In the proof, we use the Gr\"obner basis as a tool of computational algebra. 
See for example \cite{cox2013ideals} about definition and application of the Gr\"obner basis.
\\

\Proof
(a) Since the maximum number of possible values of $(t_1,\dots,t_n)$ does not increase with $n$, it is enough to prove for $n=3$. 
We set $\theta_1,\theta_2,\theta_3\in [0,\pi]$ and $a_1,a_2,a_3\geq 0$ as (\ref{eq:angles}).
Then by the law of cosine,
\begin{align}
t_2^2+t_3^2-2t_2 t_3 \cos \theta_1 &= a_1^2, \nonumber\\
t_3^2+t_1^2-2t_3 t_1 \cos \theta_2 &= a_2^2, \label{eq:a2}\\
t_1^2+t_2^2-2t_1 t_2 \cos \theta_3 &= a_3^2. \nonumber
\end{align}
We consider this as a system of polynomial equations with variables $t_1,t_2,t_3$
and compute the Gr\"obner basis of the ideal generated by the corresponding polynomials by degree lexicographic monomial order (deglex) with $t_1>t_2>t_3$
by Mathematica. 
Then the output becomes as in Note \ref{note:1} and the basis includes
$-t_1^2+(2\cos\theta_2)t_1 t_3-t_3^2+a_2^2$, $-t_2^2+(2\cos\theta_3)t_2 t_3-t_3^2+a_3^2$ and
$4v(\theta_1,\theta_2,\theta_3) t_3^4+
(\mbox{terms of degree$\leq 2$})$
where
\begin{equation}
v(\theta_1,\theta_2,\theta_3):=1+2\cos\theta_1\cos\theta_2\cos\theta_3-\cos\theta_1^2-\cos\theta_2^2-\cos\theta_3^2.
\end{equation}
Note that when $\theta_1+\theta_2+\theta_3\leq 2\pi$, $\frac{a_1 a_2 a_3}{6}v(\theta_1,\theta_2,\theta_3)$ is a formula of the volume of
the tetrahedron whose base triangle is $\triangle x_1 x_2 x_3$ and, therefore, it has a positive value unless the tetrahedron degenerates.
By the assumption, $z_1,z_2,z_3$ are not aligned in a geodesic and therefore the tetrahedron does not degenerate and $v(\theta_1,\theta_2,\theta_3)$ must be nonzero. Note that this becomes negative when  $\theta_1+\theta_2+\theta_3> 2\pi$.

On the other hand, it is known that the system of polynomial equations with a Gr\"obner basis $G$ has a finite number of (complex) solutions if and only if,
for each variable $x$, G contains a polynomial with a leading monomial that is a power of $x$.
Now all variables $t_1$,$t_2$ and $t_3$ satisfy such property, thus we conclude there are at most a finite number of solutions.

Then by B\'ezout's theorem, the number of solutions is at most the product of the degree of three polynomial equations, i.e. $2\times 2 \times 2=8$.
But if $(t_1,t_2,t_3)$ is a solution,  $(-t_1,-t_2,-t_3)$ is also a solution, and only one of each pair can satisfy $t_1,t_2,t_3\leq 0$.
Thus we conclude the number of possible values of $(t_1,t_2,t_3)$ is at most four.
\vspace{2mm}
 
(b) By the assumptions in (a), without loss of generality, we can assume $z_1$, $z_2$, $z_3$ are not aligned in a geodesic. 
By the result of (a), given $z_1$, $z_2$, $z_3$ and distances $\tilde{d}_\beta(x_1,x_2)$, $\tilde{d}_\beta(x_1,x_3)$,$\tilde{d}_\beta(x_2,x_3)$,
there are at most four variations of the values of $(t_1,t_2,t_3)$.
Here we assume $t_1$ can take multiple values including $\hat{t}_1$ and $\check{t}_1$.

Suppose, in addition to the above, the values of $z_4$ and $\tilde{d}_\beta(x_1,x_4)(=:a_4)$ are given and let $\theta_4:=\pi \min(d_Z(z_1,z_4)/\beta,1)$. Then both $\hat{t}_1$ and $\check{t}_1$ satisfy
$t_1^2+t_4^2-2t_1 t_4 \cos \theta_4 = a_4^2$
and therefore $2t_4\cos\theta_4=\hat{t}_1+\check{t}_1$ must hold.
Since $\hat{t}_1$ and $\check{t}_1$ are different non-negative values, $\hat{t}_1+\check{t}_1>0$ and, therefore, $\cos\theta_4\neq 0$.
Hence we obtain $t_4=(\hat{t}_1+\check{t}_1)/2\cos\theta$.

This means if $t_4$ takes values except $(\hat{t}_1+\check{t}_1)/2\cos\theta$, at most only one of $\hat{t}_1$ and $\check{t}_1$ can be a solution.
We can reduce each pairwise ambiguity of the (at most) 4 possibilities of $(t_1,t_2,t_3)$ one by one similarly.
Finally $(t_1,t_2,t_3)$ are determined uniquely for all except at most ${4 \choose 2} = 6$ values of $t_4$.
But such finite values of $t_4$ can be neglected thanks to the assumption of ``general'' position in the theorem.
Since the same argument holds for any triplets, the statement has been proved.
\vspace{2mm}

(c) If $(t_1,\dots,t_n)$ can take multiple values, without loss of generality we can assume $(t_1,t_2,t_3)$ takes multiple values.
By the assumption in the theorem, $\theta_1, \theta_2, \theta_3\geq \pi/2$ and therefore all coefficients in each equation of (\ref{eq:a2}) become positive.
Thus, if $t_i$ increases/decreases then $t_j$ must decrease/increase for $(i,j)=(1,2),(2,3),(3,1)$ but this cannot happen simultaneously. 
Hence $(t_1,t_2,t_3)$ cannot take multiple values.

Note that all of this proof works even when $\theta_1+\theta_2+\theta_3$ is larger than $2\pi$. \qed

\begin{remark}
Assumption in Theorem \ref{thm:a1} (a) is necessary.
If the assumption fails, the tetrahedron degenerates and $x_1,x_2,x_3$ and the apex $O$ are all in a plane.
When $O$ happens to be on a circle passing through $x_1,x_2$ and $x_3$, 
move $O$ to another point $O'$ on the same circle.
Then the angles corresponding to $\theta_1,\theta_2,\theta_3$ do not change by the inscribed angle theorem.
By an elemental geometrical argument, a new position of $x_1,x_2,x_3$ and $O'$ gives another solution of $t_1,t_2,t_3$.
Hence obviously there are infinite number of solutions.
\end{remark}

\begin{remark}
The assumption of  ``general''  positions of $z_1,\dots,z_n$ in Theorem \ref{thm:a1} (b)
is satisfied easily for most data distributions.
For example, if both $z_1,\dots,z_n\in \mathbb{R}^d$ and $t_1,\dots,t_n\in [0,1]$ are i.i.d. from a probability distribution whose density function exists with respect to the Lebesgue measure, then it is easy to see the assumption holds almost surely and
therefore uniqueness of the solution is guaranteed.
Note that for $n=3$ under the same setting,  there can be multiple solutions with a positive probability.
\end{remark}

\begin{note} \label{note:1}
Computation of the Gr\"obner basis by Mathematica:

For simplicity, we put $x:=t_1$, $y:=t_2$, $z:=t_3$, $a:=2\cos\theta_1$, $b:=2\cos\theta_2$, $c:=2\cos\theta_3$,
$d:=a_1^2$, $e:=a_2^2$ and $f:=a_3^2$.

Note that the second, first and last polynomials in the output correspond to 
$-t_1^2+(2\cos\theta_2)t_1 t_3-t_3^2+a_2^2$, $-t_2^2+(2\cos\theta_3)t_2 t_3-t_3^2+a_3^2$ and
$4v(\theta_1,\theta_2,\theta_3) t_3^4+(\mbox{terms of degree $\leq 2$})$ in the proof, respectively.

\onecolumn
{\small
\begin{verbatim}

--------------------------------------------------------------
In := GroebnerBasis[{x^2 + y^2 - a*x*y - d, x^2 + z^2 - b*x*z - e, 
  y^2 + z^2 - c*y*z - f}, {x, y, z}, 
  MonomialOrder -> DegreeLexicographic]

Out = {f - y^2 + c y z - z^2, e - x^2 + b x z - z^2, 
 d - x^2 + a x y - y^2, 
 d x - e x + a e y - x y^2 - b d z + b y^2 z + x z^2 - 
  a y z^2, -c d x + c e x - a c e y + b f y + c x y^2 - b y^3 + 
  b c d z - a f z + a y^2 z - c x z^2 - b y z^2 + a z^3, 
 a f x + d y - f y - x^2 y - c d z + c x^2 z - a x z^2 + y z^2, 
 b f x - c e y + c x^2 y - b x y^2 + e z - f z - x^2 z + 
  y^2 z, -c e x + a b f x + c x^3 + b d y - b f y - b x^2 y - 
  b c d z + a e z - a x^2 z + c x z^2 + b y z^2 - a z^3, 
 a c d x - a c e x + b f x + c d y - c e y + a^2 c e y - a b f y - 
  b x y^2 + a b y^3 - c y^3 - a b c d z + e z - f z + a^2 f z - 
  x^2 z + y^2 z - a^2 y^2 z + a c x z^2 + a b y z^2 - 
  a^2 z^3, -a e f - d x y + e x y + f x y + c d x z - c e x z + 
  b d y z - b f y z - b c d z^2 + a e z^2 + a f z^2 - 2 x y z^2 + 
  c x z^3 + b y z^3 - a z^4, -c d e + c e^2 - a b e f + c d x^2 - 
  c e x^2 - b d x y + b e x y + b f x y - a f x z - d y z + 
  b^2 d y z + 2 e y z + f y z - b^2 f y z - x^2 y z + 2 c d z^2 - 
  b^2 c d z^2 + a b e z^2 - 2 c e z^2 + a b f z^2 + a x z^3 - 
  3 y z^3 + b^2 y z^3 - a b z^4 + c z^4, -d x + a^2 d x + a b c d x + 
  2 e x - a^2 e x - a b c e x - a^2 f x + b^2 f x - x^3 + b c d y - 
  a e y + a^3 e y - b c e y + a^2 b c e y + a f y - a b^2 f y + 
  x y^2 - b^2 x y^2 - a y^3 + a b^2 y^3 - b c y^3 + b d z - 
  a^2 b d z + a c d z - a b^2 c d z + b e z - a c e z - b f z + 
  a^2 b f z - 2 x z^2 + 2 a^2 x z^2 - a^3 y z^2 + a b^2 y z^2 - 
  a^2 b z^3 + a c z^3, -c d^2 + c d e + a b d f + c d x^2 - c e x^2 + 
  b f x y - a b d y^2 + 2 c d y^2 - 2 c e y^2 + a^2 c e y^2 - 
  a b f y^2 - b x y^3 + a b y^4 - c y^4 - a d x z + a e x z - 
  a f x z - 2 d y z + e y z - a^2 e y z - f y z + a^2 f y z + 
  x^2 y z + 3 y^3 z - a^2 y^3 z, 
 d^2 - 2 d e + c^2 d e + e^2 - c^2 e^2 - 2 d f + b^2 d f + 2 e f - 
  a^2 e f + a b c e f + f^2 - b^2 f^2 - c^2 d x^2 + c^2 e x^2 + 
  b c d x y - b c e x y - b c f x y - b^2 d y^2 + b^2 f y^2 + 
  a c d x z - a c e x z + a c f x z + a b d y z + a b e y z - 
  a b f y z + 4 d z^2 - 2 b^2 d z^2 - a b c d z^2 - 2 c^2 d z^2 + 
  b^2 c^2 d z^2 - 4 e z^2 + a^2 e z^2 - a b c e z^2 + 2 c^2 e z^2 - 
  4 f z^2 + a^2 f z^2 + 2 b^2 f z^2 - a b c f z^2 + 4 z^4 - a^2 z^4 - 
  b^2 z^4 + a b c z^4 - c^2 z^4}
\end{verbatim}
}
\end{note}

\end{document}